\documentclass{article}

\usepackage[preprint]{corl_2026} 
\usepackage{etoolbox}

\makeatletter
\patchcmd{\@maketitle}{\bf \@title\par}{\bf \@title\par\vskip 0.18in\relax}{}{}
\makeatother

\usepackage{amsmath}
\usepackage{amssymb}
\usepackage{bm}
\usepackage{graphicx}
\usepackage{caption}
\usepackage{subcaption}
\usepackage{enumitem}
\usepackage{booktabs}
\usepackage{multirow}
\usepackage{wrapfig}
\usepackage{xcolor}
\usepackage{pifont}
\usepackage{etoc}
\usepackage{tcolorbox}
\usepackage{hyperref}

\newcommand{\cmark}{\textcolor{green!60!black}{\ding{51}}}
\newcommand{\xmark}{\textcolor{red!80!black}{\ding{55}}}

\title{QuadVerse: An Integrated Framework Aligning  Visual-Physical Reality for Quadruped Simulation}

\author{
\normalfont\mdseries
\begin{tabular}{c}
Yuxiang Chen$^{1,*}$ \quad
Yuanhao Wang$^{2,*}$ \quad
Ziheng Zhang$^{3,\dagger}$ \quad
Meng Zhang$^{3,\dagger}$ \\
Yu Liu$^{3}$ \quad
Yufei Jia$^{4}$ \quad
Tiancai Wang$^{3}$ \quad
Erjin Zhou$^{3}$ \quad
Jin Xie$^{1,\ddagger}$ \\[3mm]
$^{1}$Nanjing University \quad
$^{2}$BUPT \quad
$^{3}$DEXMAL \quad
$^{4}$Tsinghua University \\[1mm]
{\small $^{*}$Equal contribution \quad
$^{\dagger}$Project lead \quad
$^{\ddagger}$Corresponding author}
\end{tabular}
}

\begin{document}

\maketitle

\vspace{-2.7\baselineskip}
\begin{center}
  \includegraphics[width=0.98\textwidth]{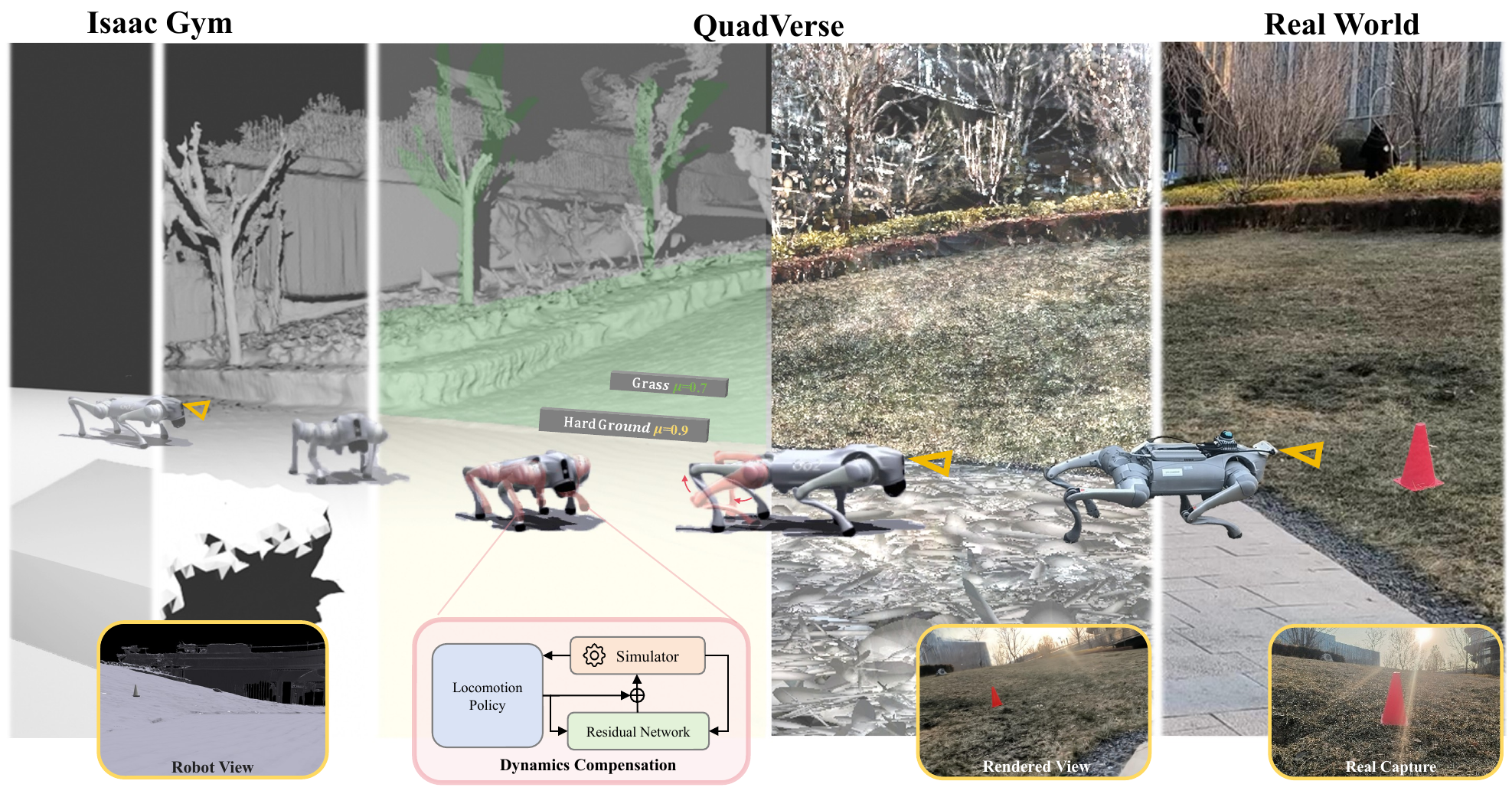}
  \captionof{figure}{
    \textbf{QuadVerse} augments existing physics simulators with batched photorealistic ego-view rendering, 
    semantic mesh-based contact calibration, 
    and an in-situ residual dynamics compensator. 
    Together, these components reduce sim-to-real discrepancies across 
    \textbf{visual perception}, \textbf{physical interaction}, and \textbf{actuator dynamics}, 
    enabling zero-shot visual-navigation policy deployment.Project page: \url{https://quad-verse.github.io/}.
  }
  \label{fig:teaser}

\end{center}


\begin{abstract}
    Simulation is central to robot learning, yet the sim-to-real gap remains a major bottleneck. 
    Existing approaches often tackle visual or dynamic gaps separately, overlooking how these individual mismatches accumulate and propagate throughout the robot's state evolution. 
    In this paper, we introduce \textbf{QuadVerse}, an integrated framework that uses reconstructed scenes as a calibration substrate for aligning visual perception, physical interaction, and actuator dynamics.
    From captured RGB videos, we reconstruct geometry-constrained 3D Gaussian Splatting (3DGS) scenes that support batched photorealistic ego-view rendering and collision-ready semantic mesh extraction. The meshes further enable contact calibration by initializing spatially varying friction priors and refining them through trajectory-based posterior search. 
    To address remaining actuator discrepancies, QuadVerse trains a residual dynamics compensator by replaying real-world trajectories on the contact-calibrated terrain, reducing the entanglement between terrain-induced contact errors and actuator non-idealities. 
    Experiments show that QuadVerse improves reconstruction quality and locomotion tracking over relevant baselines. 
    Leveraging this foundation, we demonstrate robust zero-shot visual-navigation policy deployment without task-specific real-world rollouts.
\end{abstract}



\section{Introduction}

Legged robots, particularly quadrupeds, have demonstrated exceptional mobility in complex, unstructured terrains~\cite{bledt2018cheetah, katz2019mini, lee2020learning}. Deep Reinforcement Learning (DRL) \cite{hwangbo2019learning, miki2022learning} has emerged as the prevailing paradigm for learning agile locomotion skills. Due to hardware costs and safety risks, physics-based simulators have become the cornerstone of policy training \cite{makoviychuk2021isaac, mittal2025isaac, todorov2012mujoco, zakka2025mujoco}. However, the sim-to-real gap remains a major bottleneck for transferring policies to the real world~\cite{zhao2020sim}. 

Recent work has made substantial progress on individual aspects of sim-to-real transfer. 
Real-to-sim reconstruction and 3DGS-based simulators improve visual realism for vision-based policy learning~\cite{xie2024vid2sim, zhu2025vr, chhablani2025embodiedsplat, escontrela2025gaussgym}, while domain randomization and system identification improve robustness to dynamics mismatch~\cite{tobin2017domain, peng2018sim, siekmann2021sim, masuda2023sim, o2022neural, sobanbabu2025sampling}.
For contact-rich quadruped deployment, however, these discrepancies interact: visual observations determine policy actions, terrain geometry and contact parameters shape external forces, and actuator delays or nonlinearities alter the executed motion. 
Errors at any of these interfaces accumulate into trajectory mismatch, motivating an integrated framework that systematically reduces these discrepancies. 

Guided by this perspective, we present \textbf{QuadVerse}, an integrated framework that uses reconstructed scenes as a shared substrate for aligning visual perception, physical interaction, and actuator dynamics. 
From captured RGB videos, QuadVerse reconstructs geometry-constrained 3D Gaussian Splatting (3DGS) \cite{kerbl3Dgaussians} scenes that support batched photorealistic ego-view rendering and collision-ready semantic mesh extraction. 
The resulting meshes enable spatially varying contact calibration, where coarse friction priors are initialized from terrain semantics and refined through trajectory-based posterior search. 
To reduce the remaining actuator mismatch, QuadVerse trains a residual actuator dynamics compensator by replaying real-world trajectories on the contact-calibrated terrain, thereby reducing the entanglement between terrain-induced contact errors and actuator non-idealities.

We validate QuadVerse through targeted experiments that evaluate both module-level fidelity and system-level sim-to-real transfer. 
Our reconstruction pipeline achieves accurate geometry and high-fidelity rendering while supporting high-throughput ego-view generation at over 2000 FPS at \(640\times480\) resolution for RL training.
Our prior-posterior contact calibration reduces mean position error during command replay on mixed-friction terrain, and our residual actuator compensation improves joint-space replay accuracy and downstream locomotion deployment.
Finally, in an integrated outdoor navigation task, zero-shot deployment of policies trained in QuadVerse attains an 84\% real-world success rate, compared with 92\% in the aligned simulation, without task-specific real-world rollouts.
We will release the code to facilitate future research.


\section{Related Works}
\label{sec:related_works}

\textbf{Visual Sim-to-Real Adaptation}\hspace{1em}High-fidelity visual simulation is essential for minimizing domain shifts in vision-based policy training.
While conventional rasterization-based simulators require substantial manual asset design to reproduce complex scenes, 3D Gaussian Splatting (3DGS)~\cite{kerbl3Dgaussians} has recently emerged as an effective representation for constructing photorealistic robot simulation environments, including manipulation~\cite{lou2025robo, li2024robogsim, jia2025discoverse} and navigation~\cite{zhu2025vr, xie2024vid2sim, chhablani2025embodiedsplat, escontrela2025gaussgym} tasks.
These works leverage 3DGS to construct photorealistic digital twins, significantly narrowing the visual gap. 
However, existing 3DGS-based real-to-sim systems primarily use reconstructed scenes for visual rendering or geometric interaction, while their role in calibrating contact properties and actuator responses remains underexplored. 
In contrast, QuadVerse uses reconstructed 3DGS scenes as a geometric substrate for contact calibration and residual dynamics compensation, while also supporting scalable batched ego-view rendering for RL training.

\textbf{Terrain Contact Parameter Alignment}\hspace{1em}
Inferring exact terrain contact parameters solely from visual appearance is inherently ill-posed, as visually similar materials may differ significantly in properties.
Prior works have used semantic terrain understanding, learned vision-to-physical-parameter prediction, or VLM/LLM-based physical priors to estimate terrain attributes before contact~\cite{ewen2023multi, chen2024identifying, peng2024friction, xu2025gaussianproperty}. 
These approaches provide useful priors, but directly using semantic labels or appearance-based predictions as simulator parameters remains fragile due to intra-class variance. 
Interaction-based and system-identification methods can further refine terrain parameters from real robot responses, but often require dedicated probing motions, online estimation, or active exploration~\cite{margolis2023learning, yu2017preparing, kim2025online, sobanbabu2025sampling}. 
QuadVerse therefore adopts a lightweight two-stage strategy: 3D semantics initialize coarse spatially varying friction priors, while trajectory-based posterior search refines contact parameters by replaying real-world locomotion on the contact-calibrated terrain.

\textbf{Dynamics Sim-to-Real Alignment}\hspace{1em}
Dynamics mismatch remains a persistent bottleneck for sim-to-real robot learning. 
Domain randomization (DR) improves robustness by perturbing physical parameters during training~\cite{tobin2017domain, peng2018sim, siekmann2021sim}, but may trade tracking accuracy for conservative behavior. 
System identification (SysID) instead refines simulator parameters from real-world data~\cite{tan2018sim, masuda2023sim, o2022neural, sobanbabu2025sampling}, yet remains limited by the expressiveness of predefined analytical models.
Beyond such parameter-level alignment, data-driven approaches can further model residual dynamics or actuator responses that remain unexplained by nominal simulators.
Prior works have learned neural actuator models for legged locomotion~\cite{hwangbo2019learning}, compensators from matched command-response data~\cite{fey2025bridging}, neural dynamics models for dexterous manipulation~\cite{liu2025dexndm}, or full-body residual models using rich motion-capture supervision~\cite{he2025asap}. 
However, low-cost residual compensation for high-load floating-base robots remains challenging, especially when terrain-induced contact errors can be entangled with actuator-level discrepancies. 


\section{Problem Formulation}
\label{sec:problem}

We view sim-to-real transfer as minimizing the discrepancy between simulated and real state trajectories. 
Let $\boldsymbol{q}$ and $\boldsymbol{v}$ denote the generalized coordinates and velocities of the quadruped. 
The constrained rigid-body dynamics can be written as
\begin{equation}
    \boldsymbol{M}(\boldsymbol{q})\dot{\boldsymbol{v}} 
    + \boldsymbol{C}(\boldsymbol{q},\boldsymbol{v}) 
    + \boldsymbol{G}(\boldsymbol{q}) 
    = \boldsymbol{\tau} + \boldsymbol{J}^T \boldsymbol{f}_{ext},
    \label{eq:dynamics}
\end{equation}
where $\boldsymbol{M}$, $\boldsymbol{C}$, and $\boldsymbol{G}$ denote the inertia, Coriolis/centrifugal, and gravity terms, $\boldsymbol{\tau}$ is the joint torque, $\boldsymbol{f}_{ext}$ is the external contact force, and $\boldsymbol{J}$ is the contact Jacobian. 
After nominal inertial calibration, the dominant trajectory mismatch can be organized around three interfaces: visual observations that determine policy actions, contact modeling that determines $\boldsymbol{f}_{ext}$, and actuator response that determines the executed torque $\boldsymbol{\tau}$.
    
In vision-based control, the commanded torque is generated through a policy and an actuator model:
\begin{equation}
    \boldsymbol{\tau}_t
    = \mathcal{A}(\boldsymbol{a}_t)
    = \mathcal{A}\big(\pi_\theta(\boldsymbol{s}_t,\boldsymbol{I}_t)\big),
    \label{eq:policy_torque}
\end{equation}
where $\boldsymbol{s}_t$ is the proprioceptive state and $\boldsymbol{I}_t$ is the visual observation. 
A domain shift between simulated and real images changes the policy input distribution and can therefore induce mismatched actions. 
QuadVerse reduces this discrepancy with high-fidelity 3DGS rendering.

The physical interaction gap appears in the external force term. 
We write it abstractly as
\begin{equation}
    \boldsymbol{f}_{ext} = \mathcal{F}_{contact}(\boldsymbol{v}_{rel}, \boldsymbol{n}, \boldsymbol{\Phi}),
    \label{eq:contact_force}
\end{equation}
where $\boldsymbol{v}_{rel}$ is the relative contact velocity, $\boldsymbol{n}$ is the contact normal, and $\boldsymbol{\Phi}$ denotes contact parameters such as friction. 
Coarse terrain geometry or uniform contact parameters can bias both $\boldsymbol{n}$ and $\boldsymbol{\Phi}$. 
QuadVerse therefore extracts collision-ready meshes to improve geometric contact modeling and uses semantic contact calibration to assign spatially varying contact parameters.

Finally, even with aligned observations and contact modeling, the simulated actuator model $\mathcal{A}_{sim}$ may differ from the real actuator response $\mathcal{A}_{real}$ due to delays, nonlinear friction, and other motor non-idealities. 
We model the remaining effective actuation mismatch as a residual term:
\begin{equation}
    \boldsymbol{\tau}_{real} 
    = \boldsymbol{\tau}_{sim} 
    + \Delta_{act}(\boldsymbol{s}_t,\boldsymbol{a}_t),
    \label{eq:actuator_residual}
\end{equation}
where $\boldsymbol{\tau}_{sim}$ denotes the torque predicted by the nominal actuator model, $\boldsymbol{\tau}_{real}$ denotes the torque executed by the real robot, and $\Delta_{act}$ captures actuator-level discrepancies.
In practice, this residual is implemented in the action space. 


\section{Method}
\label{sec:method}

\begin{figure*}[t]
    \vspace{-2\baselineskip}
    \centering
    \includegraphics[width=0.95\linewidth]{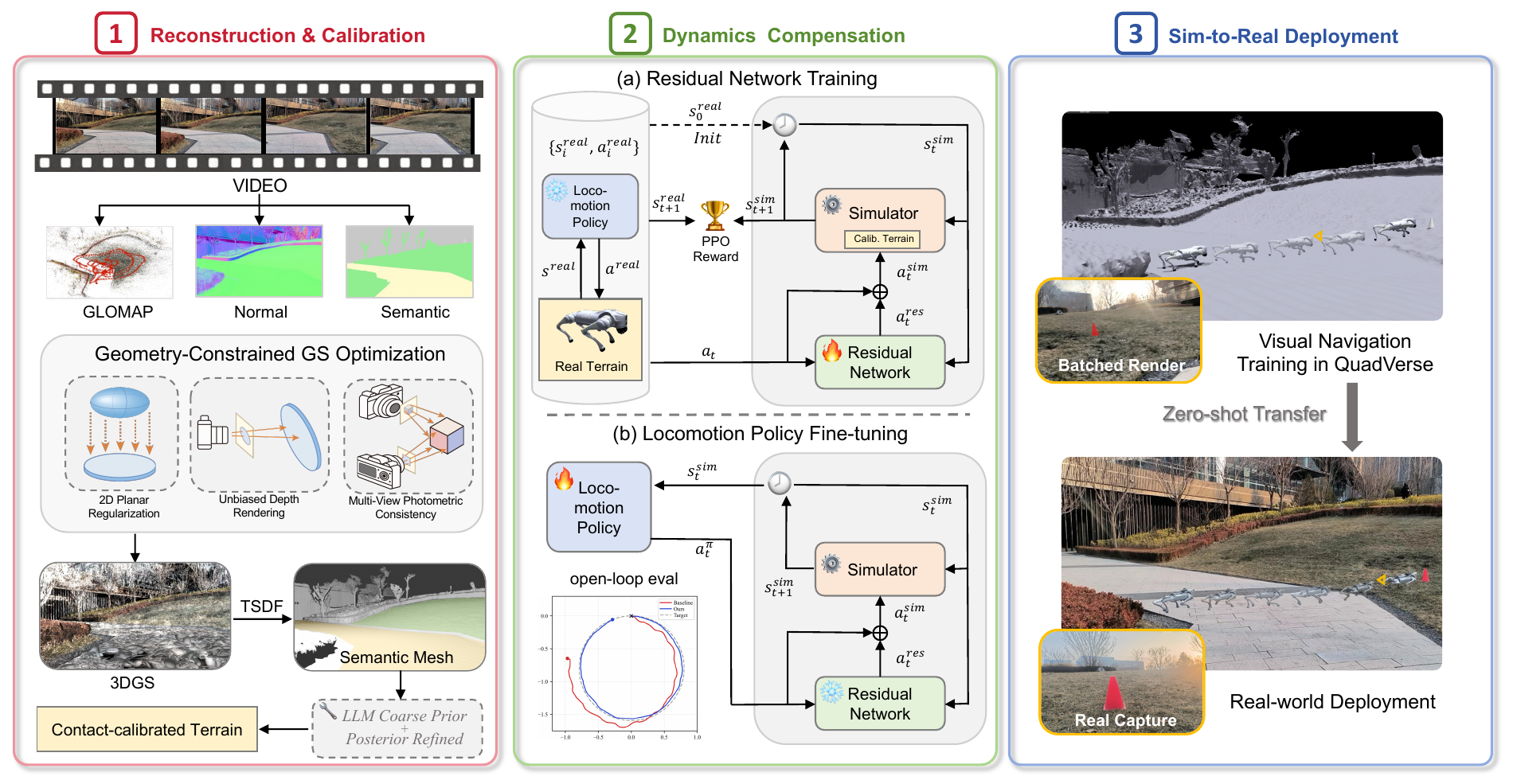}
    \vspace{-0.5\baselineskip}
    \caption{Overview of QuadVerse. (1) \textbf{Reconstruction and Calibration:} QuadVerse reconstructs 3DGS scenes for batched ego-view rendering and extracts collision-ready semantic meshes for contact calibration.  (2) \textbf{Dynamics Compensation:} A dynamics compensator is trained using RL by replaying real-world trajectories on the contact-calibrated terrain; the locomotion policy is then fine-tuned under the corrected dynamics.  (3) \textbf{Sim-to-Real Deployment:} Policies trained in QuadVerse are deployed zero-shot to outdoor visual navigation tasks without task-specific real-world rollouts.}
    \label{fig:pipeline}
    \vspace{-1.2\baselineskip}
\end{figure*}

Building upon the perspective outlined above, \textbf{QuadVerse} aims to achieve reliable sim-to-real transfer by coordinating visual perception, physical interaction, and actuator dynamics within a reconstructed real-to-sim environment, as shown in Fig.~\ref{fig:pipeline}. 

\subsection{Geometry-Anchored Reconstruction and Batched Rendering}

\textbf{Geometry-Consistent Scene Reconstruction.} To create photorealistic and geometrically accurate digital twins, we reconstruct 3D scenes from captured RGB video sequences. We initialize camera poses and sparse geometry via global structure-from-motion (GLOMAP \cite{pan2024global}), which effectively prevents cumulative drift in large outdoor scenes.
The scene is then represented with 3D Gaussian Splatting (3DGS) \cite{kerbl3Dgaussians}. 
However, vanilla 3DGS is primarily optimized for photometric reconstruction and may produce inaccurate surface geometry, which are unsuitable for collision modeling.

To extract collision-ready meshes for downstream physical simulation, we introduce geometric constraints during 3DGS optimization. 
Specifically, we flatten Gaussian primitives via 2D planar regularization~\cite{huang20242dgs}, reduce depth bias through unbiased depth rendering~\cite{chen2024pgsr}, and enforce multi-view photometric consistency using induced homographies~\cite{campbell2008using, fu2022geo, chen2024pgsr}. 
For texture-less regions where photometric supervision is weak, we incorporate normal priors predicted by StableNormal~\cite{ye2024stablenormal} to regularize the rendered surface normals. 
Detailed formulations are provided in the Appendix.

\textbf{High-Throughput Batched Rasterization.} 
To support visual RL training with many parallel environments, QuadVerse implements batched ego-view rendering directly within the simulation pipeline. 
Instead of rendering each environment sequentially, we batch camera matrices across parallel environments while sharing static scene Gaussians. 
Dynamic Gaussian groups, such as task objects, are transformed according to their per-environment poses. 
The resulting batched scene representation is rendered using a tile-based Gaussian rasterizer~\cite{ye2025gsplat, jia2026gs}, enabling high-throughput ego-view generation for policy training.

\subsection{Semantic Mesh-Based Contact Calibration}

To reduce the physical interaction gap, QuadVerse converts the reconstructed 3DGS scene into a collision-ready semantic mesh and calibrates spatially varying contact parameters on top of it. 

\textbf{Semantic Collision Mesh Extraction.} 
After 3DGS optimization, unbiased depth maps are rendered from all training views and fused into a volumetric Truncated Signed Distance Function (TSDF)~\cite{newcombe2011kinectfusion}. 
A watertight mesh is extracted via Marching Cubes~\cite{lorensen1998marching}. 
Metric scale and coordinate alignment are obtained through a similarity transformation based on known scene markers, and the mesh is transformed into the simulator's $z$-up frame for physics simulation.

To associate contact properties with terrain regions, we augment each Gaussian primitive with a learnable semantic feature supervised by pseudo-labels from a 2D segmentation model~\cite{chen2022vision}. 
The learned 3D semantic features are then propagated to mesh faces via k-nearest neighbors (kNN)~\cite{cover1967nearest}, producing a semantic collision mesh that supports region-wise contact calibration.

\textbf{Prior-Posterior Friction Calibration.} 
Since the outdoor terrains considered in this work are mostly rigid and the dominant contact failure is tangential slip, we focus on friction as the primary calibrated parameter. 
For each semantic terrain region, a commonsense reasoning model (e.g., GPT-4~\cite{achiam2023gpt}) provides a coarse nominal friction prior. 
For regions with sufficiently high friction, we keep the nominal prior without posterior refinement, since the robot trajectory becomes insensitive to the exact value of $\mu$ above an empirical slip threshold, set to $\mu=0.5$ in our experiments.

For slip-prone regions, we perform a grid search over a region-level friction coefficient by replaying the recorded locomotion commands and selecting the value that minimizes trajectory discrepancy:
\begin{equation}
    \mu_r^\star = 
    \arg\min_{\mu_r \in [\mu_{\min}, \mu_{\max}]}
    \sum_{t} 
    \left\|
    \boldsymbol{x}^{sim}_{t}(\mu_r) - \boldsymbol{x}^{real}_{t}
    \right\|_2^2,
    \label{eq:friction_search}
\end{equation}
where $r$ denotes a slip-prone semantic region and $\boldsymbol{x}_t$ denotes the trajectory state used for calibration. 
Details of the calibration protocol are provided in the Appendix.

\subsection{Dynamics Compensation via Residual RL}
Even with visual and contact interfaces aligned, the nominal simulator may still fail to capture actuator-level effects such as signal delays, internal motor friction, and nonlinear response. 
To address this, QuadVerse trains an action-space residual dynamics compensator from real-world trajectory replay on the contact-calibrated terrain.

\textbf{Data Collection and Replay.}
We collect approximately 10 minutes of real-world locomotion data across the targeted unstructured terrain. This dataset comprises high-level locomotion commands, joint-space commands, high-frequency proprioceptive states, and LiDAR-tracked global base poses.
During replay, each simulation episode is initialized from a sampled timestamp using the corresponding base pose and joint state. 
We then replay the recorded joint-space commands on the contact-calibrated terrain rather than on a flat plane. 
This better matches the real replay condition and prevents terrain-induced contact errors from being absorbed into the actuator residual.

\textbf{Residual Network Training.} 
We train a residual policy via PPO~\cite{schulman2017proximal} to output an additive action correction $\boldsymbol{a}^{res}$. 
During replay, the applied simulation action is $\boldsymbol{a}^{sim}=\boldsymbol{a}^{ref}+\boldsymbol{a}^{res}$, where $\boldsymbol{a}^{ref}$ denotes the recorded action. 
We use an asymmetric actor-critic architecture: the actor observes only simulation-available quantities, including joint states, contact states, and base pose, while the critic additionally uses privileged real-world trajectory information to guide value estimation.
The reward balances joint-space tracking, pose/contact consistency, and residual regularization. 
The tracking terms encourage the simulated replay to follow the recorded trajectory, while the regularization terms penalize large or rapidly varying residual actions.
Episodes are terminated when joint errors exceed a safety threshold, preventing training from diverging far outside the replay distribution.

\textbf{Locomotion Policy Fine-tuning.}
After training, the residual compensator is frozen and inserted into the simulation loop for downstream locomotion policy fine-tuning. 
Given a nominal policy action $\boldsymbol{a}^{\pi}$, the executed simulation action is
\begin{equation}
    \boldsymbol{a}^{sim} = \boldsymbol{a}^{\pi} + \mathrm{clip}(\boldsymbol{a}^{res}, -\boldsymbol{\delta}, \boldsymbol{\delta}),
\end{equation}
where $\boldsymbol{\delta}$ bounds the compensation magnitude. 
This clipping prevents aggressive over-compensation when the policy explores states outside the replay distribution. 
Detailed network architectures, reward terms, and hyperparameters are provided in the Appendix.

\section{Experiments}
\label{sec:result}

\subsection{Experimental Setup}
All real-world experiments are conducted on a Unitree Go2 quadruped robot. 
For visual perception, a head-mounted Intel RealSense D435i camera provides \(640\times480\) RGB images. 
A Livox Mid-360 LiDAR coupled with a FAST-LIO SLAM backend~\cite{xu2021fast} is used to collect geometric reference scans for reconstruction evaluation and to track robot base poses during replay experiments.

QuadVerse uses Isaac Gym~\cite{makoviychuk2021isaac} for rigid-body dynamics and contact simulation. 
During training, we instantiate 1024 parallel environments on a single NVIDIA RTX 4090 GPU. 
With batched 3DGS rendering using a minibatch size of 128, QuadVerse achieves over 2000 FPS rendering throughput at \(640\times480\) resolution. 
Consequently, this highly efficient pipeline completes the training of the residual compensation model and the fine-tuning of the locomotion policy within 1 hour, alongside the training of visual RL navigation policy within 2 hours. Additional experimental settings, implementation details, and extended results are provided in the Appendix.

\subsection{Visual Fidelity and Contact Modeling}

\begin{wraptable}{r}{0.5\textwidth}
    \vspace{-14pt}
    \centering
    \footnotesize
    \setlength{\tabcolsep}{2.5pt}
    \caption{\textbf{Rendering quality and simulation capability.} 
    Interact. denotes collision-ready interaction, and Batch denotes batched ego-view rendering for parallel RL.}
    \label{tab:rendering_and_capability}
    \resizebox{\linewidth}{!}{%
        \begin{tabular}{l|ccc|cc}
            \toprule
            \multirow{2}{*}{Methods} & \multicolumn{3}{c|}{Rendering} & \multicolumn{2}{c}{Simulation} \\
             & PSNR $\uparrow$ & SSIM $\uparrow$ & LPIPS $\downarrow$ & Interact. & Batch \\
            \midrule
            Instant-NGP  & 20.46 & 0.475 & 0.598 & \xmark & \xmark \\
            VGGT-X       & 16.75 & 0.299 & 0.538 & \xmark & \xmark \\
            3DGS         & 20.85 & 0.554 & \textbf{0.378} & \xmark & \xmark \\
            2DGS         & 20.57 & 0.524 & 0.456 & \cmark & \xmark \\
            Vid2Sim      & 20.18 & 0.533 & 0.384 & \cmark & \xmark \\
            \midrule
            \textbf{Ours} & \textbf{21.04} & \textbf{0.569} & 0.394 & \cmark & \cmark \\
            \bottomrule
        \end{tabular}%
    }
    \vspace{-10pt}
\end{wraptable}

We first evaluate QuadVerse on 10 self-collected outdoor scenes covering ramps, grass, staircases, mixed pavements, and other unstructured terrains.
As shown in Table~\ref{tab:rendering_and_capability}, QuadVerse achieves the best average PSNR (21.04) and SSIM (0.569) among representative NeRF~\cite{muller2022instant}, 3DGS-based~\cite{kerbl3Dgaussians, huang20242dgs, xie2024vid2sim}, and feed-forward reconstruction~\cite{liu2025vggt} baselines, demonstrating superior photometric reconstruction.
More importantly for robot learning, the reconstructed scenes support both collision-ready interaction and batched ego-view rendering, enabling their use in massively parallel RL rather than only offline visualization.

\begin{figure*}[t]
    \centering
    \setlength{\tabcolsep}{1pt} 
    \renewcommand{\arraystretch}{1.2}

    \begin{tabular}{ccccc}
        \includegraphics[width=0.19\linewidth]{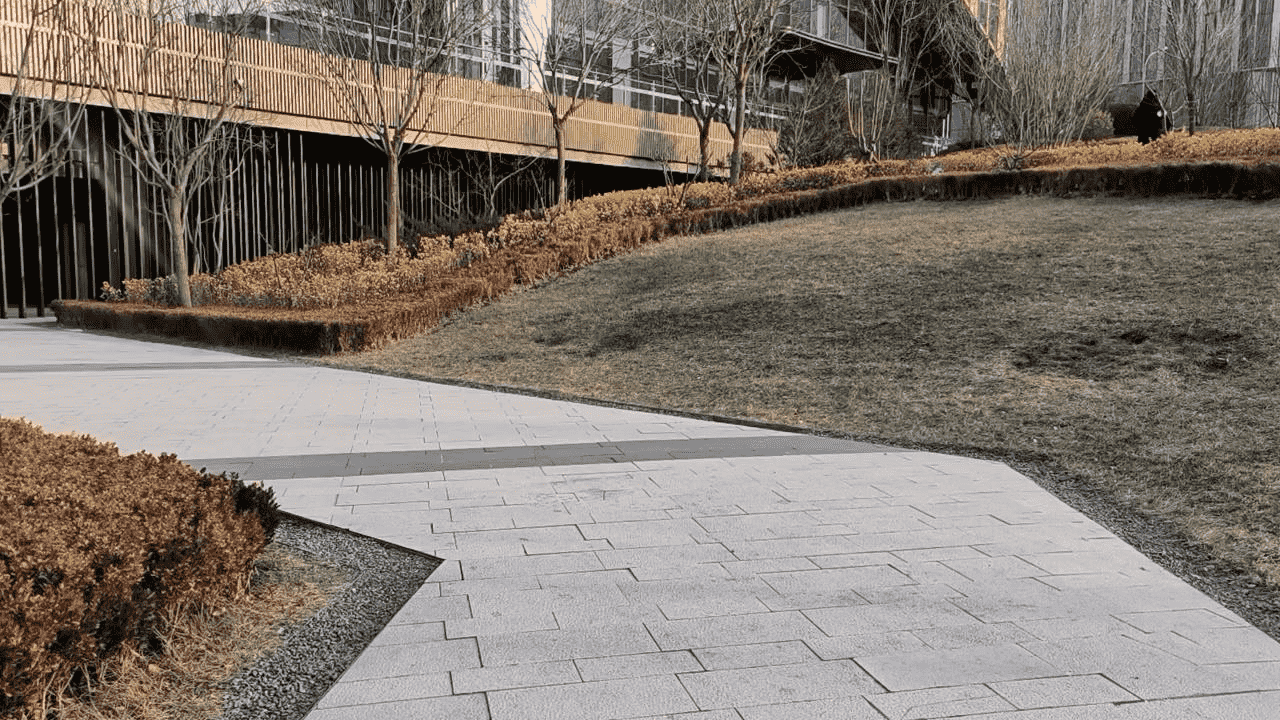} &
        \includegraphics[width=0.19\linewidth]{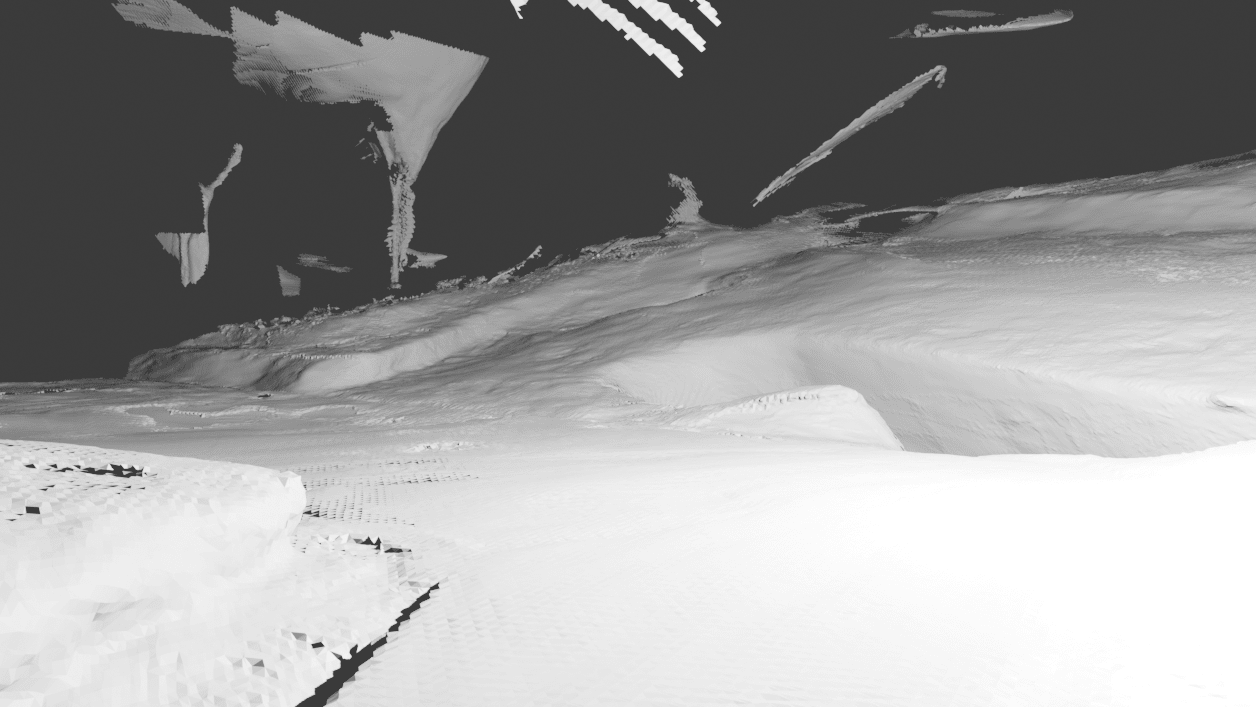} &
        \includegraphics[width=0.19\linewidth]{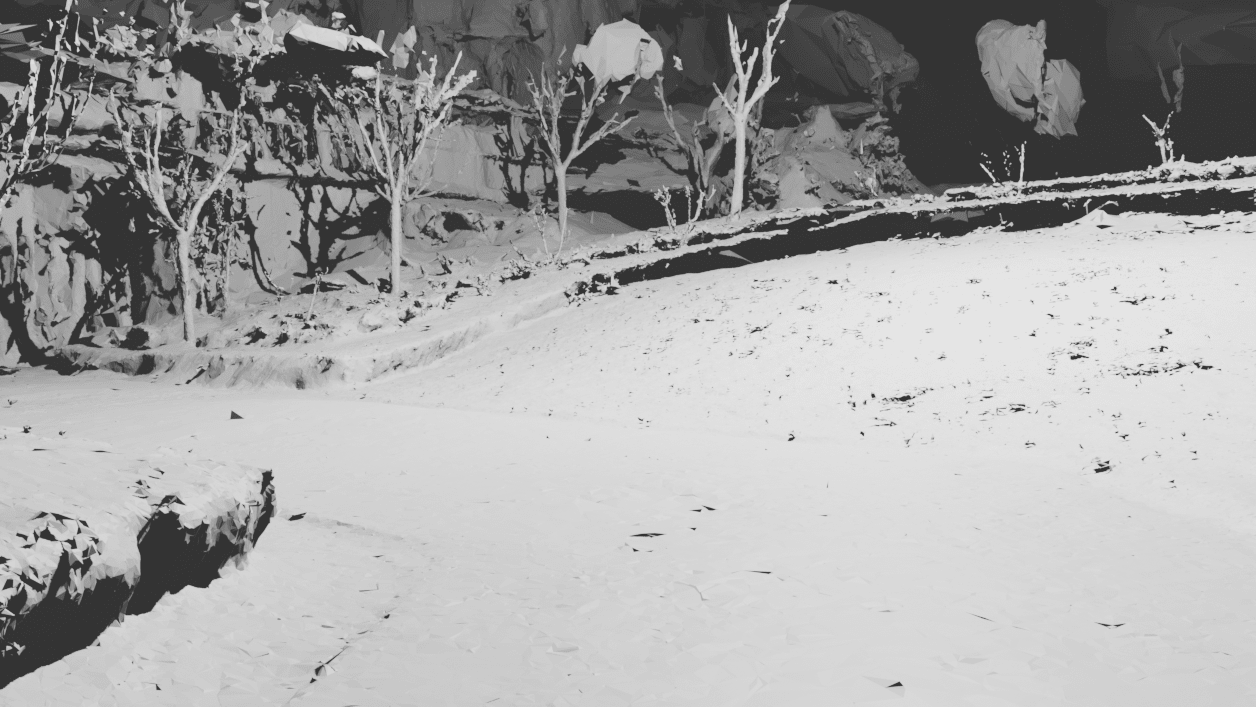} &
        \includegraphics[width=0.19\linewidth]{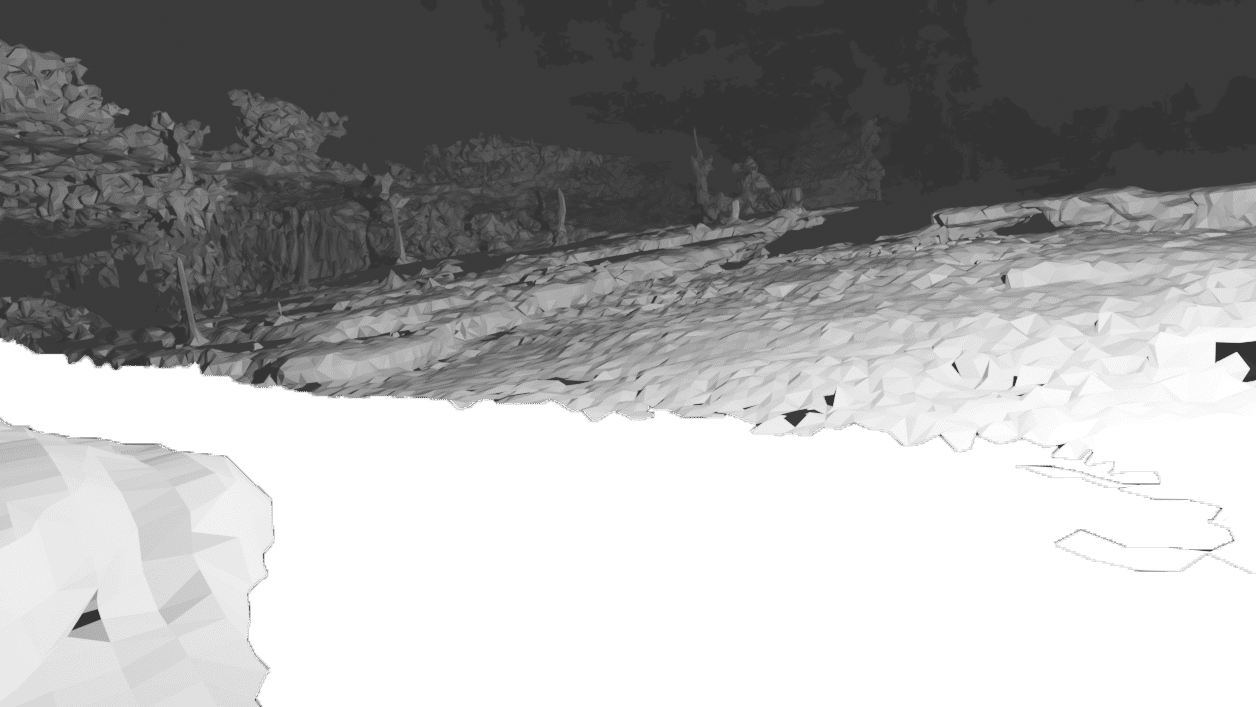} &
        \includegraphics[width=0.19\linewidth]{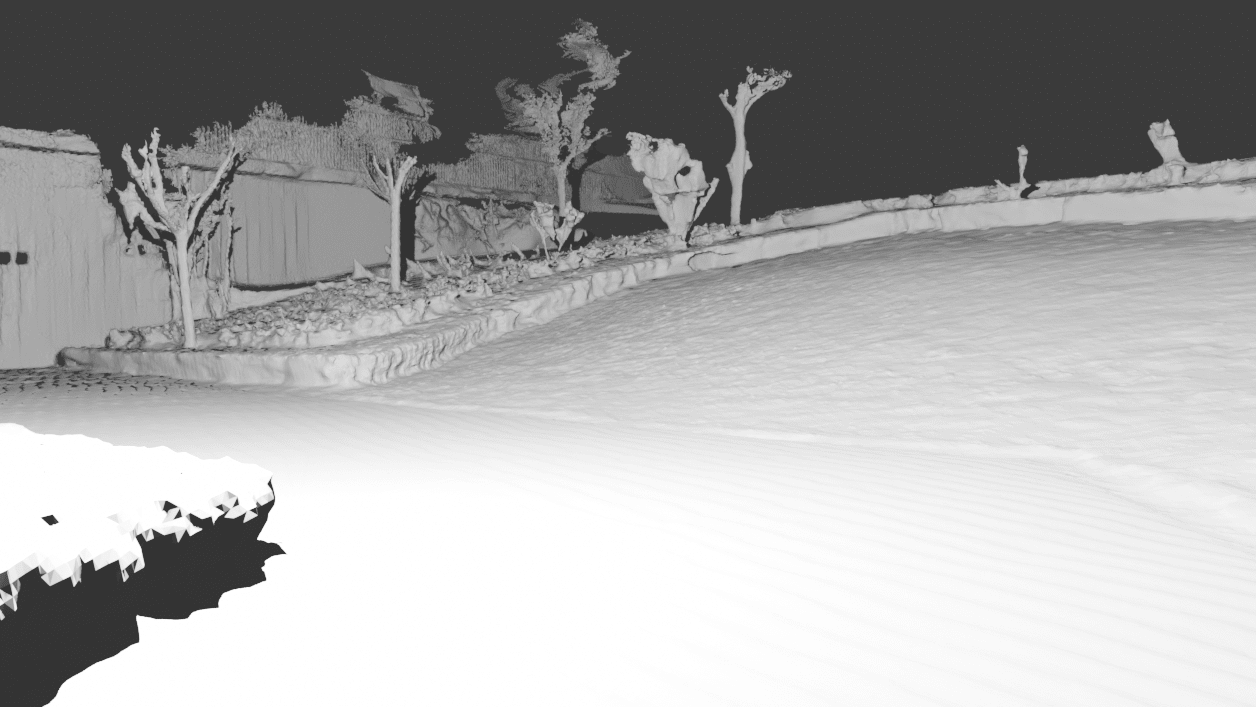} \\

        \footnotesize GT Image & 
        \footnotesize 2DGS (0.784) & 
        \footnotesize MILO (0.838) & 
        \footnotesize Vid2Sim\textsuperscript{*} (0.271)& 
        \footnotesize \textbf{QuadVerse} (0.932) \\
    \end{tabular}

    \caption{\textbf{Geometric reconstruction evaluation.} 
    We compare extracted meshes against LiDAR-scanned geometry using F1 score$\uparrow$. 
    QuadVerse reconstructs a coherent watertight mesh and achieves the highest F1 score (0.932), while competing methods suffer from over-smoothing or surface noise.
    The * indicates that Vid2Sim~\cite{xie2024vid2sim} performs ground reconstruction separately.}
    \label{fig:mesh_eval}
    \vspace{-1\baselineskip}
\end{figure*}

To balance geometric fidelity and computational efficiency during mesh extraction, we set the TSDF truncation distance to 10\,m and the voxel size to 0.01\,m.
As shown in Fig.~\ref{fig:mesh_eval}, QuadVerse reconstructs coherent watertight surfaces with sharper terrain boundaries, achieving an F1 score of 0.932 and outperforming existing geometric reconstruction baselines~\cite{huang20242dgs, guedon2025milo, xie2024vid2sim}.

\begin{wraptable}{r}{0.43\textwidth}
    \vspace{-10pt}
    \centering
    \footnotesize
    \setlength{\tabcolsep}{3pt}
    \caption{\textbf{Evaluation of contact calibration.} 
    Mean base-position error during replay on mixed-friction terrains.}
    \label{tab:slip_error}
    \vspace{-5pt}
    \resizebox{\linewidth}{!}{%
        \begin{tabular}{lc}
            \toprule
            \textbf{Contact Modeling} & \textbf{Mean Error (m) $\downarrow$} \\
            \midrule
            Uniform Friction & 0.70 \\
            LLM Coarse Prior & 0.20 \\
            \textbf{Posterior Refined (Ours)} & \textbf{0.12} \\
            \bottomrule
        \end{tabular}%
    }
    \vspace{-15pt}
\end{wraptable}

We further evaluate the effectiveness of our prior-posterior friction calibration on the mixed-friction terrains. 
Across multiple recorded replay segments, we measure the mean base-position error over time between simulated and real trajectories. 
As reported in Table~\ref{tab:slip_error}, uniform friction fails to reproduce traction loss, resulting in a large replay error. 
The LLM coarse prior, initialized from semantic terrain labels, reduces the error by identifying low-friction regions, while posterior refinement further calibrates the slip response and achieves the lowest mean error of 0.12\,m.

\subsection{Actuator Modeling and Tracking Accuracy}
We next evaluate whether the residual actuator compensator improves replay accuracy and whether the compensated dynamics benefits downstream locomotion policy fine-tuning.

\begin{wrapfigure}{r}{0.48\textwidth}
    \vspace{-10pt}
    \centering
    \includegraphics[width=\linewidth]{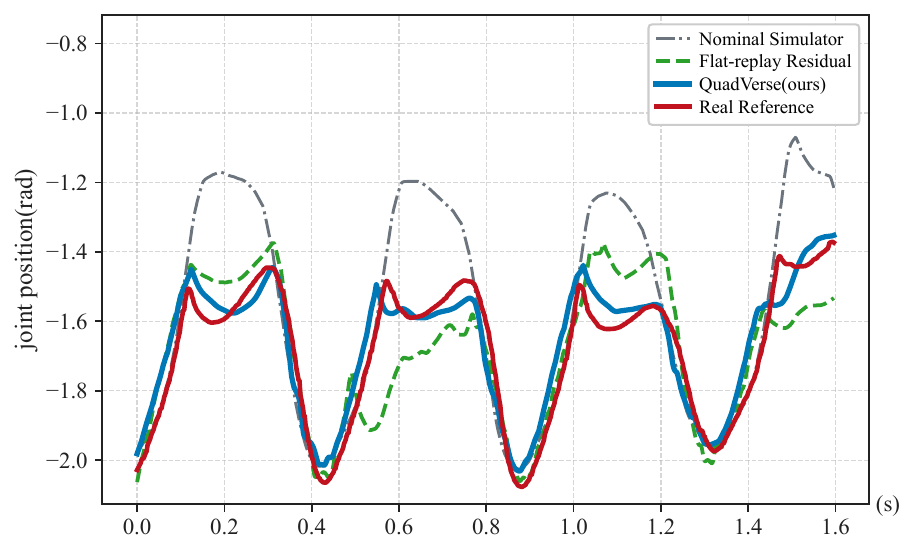}
    \caption{
    \textbf{Joint-space tracking during open-loop replay.}
    The nominal simulator shows large actuator mismatch, and flat-replay residual overcompensates under unstructured contacts. 
    QuadVerse replay on contact-calibrated terrain most closely matches the real reference.}
    \label{fig:joint_space}
    \vspace{-18pt}
\end{wrapfigure}

\textbf{Open-Loop Joint-Space Replay.}
We evaluate the residual model by replaying recorded joint-space commands and comparing simulated joint trajectories with real measurements. 
We compare three settings: \textit{Nominal Simulator}, which uses the default actuator model without compensation; \textit{Flat-Replay Residual}, where the residual model is trained from replay on flat terrain; and \textit{QuadVerse}, where the residual model is trained from replay on the contact-calibrated terrain.

As shown in Fig.~\ref{fig:joint_space}, the nominal simulator deviates substantially from real reference due to actuator latency, internal friction, and other unmodeled motor effects. 
Flat-replay residual reduces this mismatch, but remains biased. 
In contrast, QuadVerse trains the compensator under matched terrain geometry and calibrated contact conditions, achieving the lowest mean per-joint tracking error among the three settings: 0.127\,rad for Nominal Simulator, 0.064\,rad for Flat-Replay Residual, and \textbf{0.043}\,rad for QuadVerse. 
Additional experiments on Unitree Go1 and A2 are provided in the Appendix to demonstrate the applicability of our compensation method across quadruped platforms.

\begin{wrapfigure}{r}{0.48\textwidth}
    \vspace{-15pt}
    \centering
    \includegraphics[width=\linewidth]{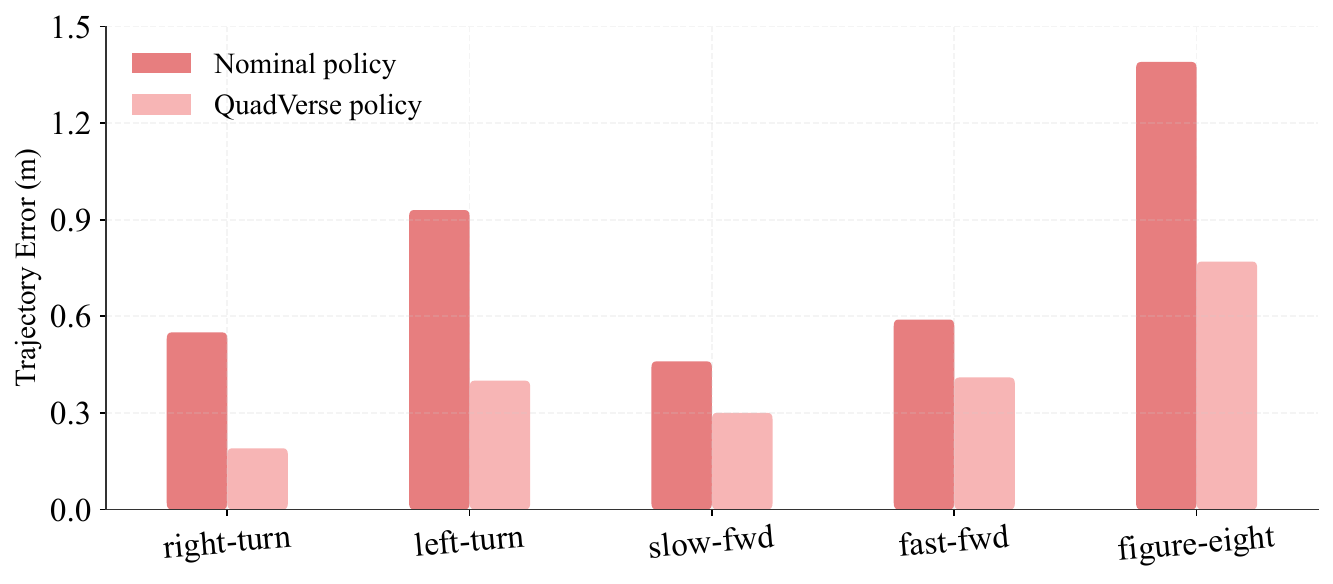}
    \caption{
    \textbf{Real-world trajectory error across locomotion tasks.}
    Policies fine-tuned with QuadVerse's residual actuator compensation achieve lower tracking errors than the nominal policy without compensation.}
    \label{fig:traj_error}
    \vspace{-18pt}
\end{wrapfigure}

\textbf{Policy Fine-Tuning and Real-World Tracking.}
After training, we freeze the residual compensator and insert it into the simulation loop for locomotion policy fine-tuning. 

\begin{figure*}[t]
    \centering
    \includegraphics[width=\linewidth]{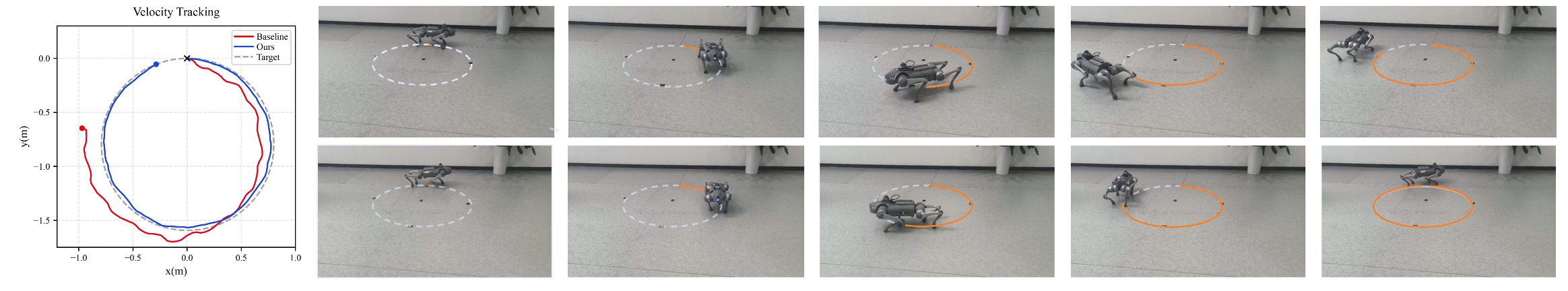}
    \caption{
    \textbf{Real-world trajectory tracking for the right-turn maneuver.}
    \textbf{Left:} Global trajectory in the world frame. 
    \textbf{Top:} The nominal policy without dynamics compensation drifts away from the reference path. 
    \textbf{Bottom:} The policy fine-tuned with QuadVerse's compensated dynamics follows the reference trajectory more closely.}
    \label{fig:real_traj}
    \vspace{-8pt}
\end{figure*}

Fig.~\ref{fig:traj_error} summarizes real-world trajectory tracking errors across multiple locomotion commands. 
Fine-tuning under the compensated dynamics reduces the average trajectory error by 47\% relative to the nominal policy, indicating that improved replay fidelity translates to better closed-loop deployment.
Fig.~\ref{fig:real_traj} provides a qualitative example on a highly dynamic right-turn maneuver. 
Without compensation, the real robot accumulates substantial drift, whereas the fine-tuned policy follows the intended trajectory more closely.

\subsection{Zero-Shot Visual Navigation Deployment}
Finally, we evaluate the full QuadVerse workflow on zero-shot real-world visual navigation. 
Following our pipeline, we reconstruct the deployment scene, calibrate the semantic contact mesh, and integrate a replay-trained residual actuator compensator before training the navigation policy. 
The robot must navigate to a colored cone on unstructured outdoor terrain using only egocentric RGB observations within a 25-second time limit.
We adopt a hierarchical RL framework~\cite{zhu2025vr}, where a 5\,Hz high-level visual planner outputs velocity commands to a 50\,Hz low-level locomotion controller. 
The visual navigation policy is trained entirely in simulation and deployed to the real robot without task-specific real-world rollouts.

\begin{wraptable}{r}{0.52\textwidth}
    \vspace{-12pt}
    \centering
    \footnotesize
    \setlength{\tabcolsep}{4pt}
    \caption{
    \textbf{Zero-shot visual navigation performance.} 
    SR denotes success rate and ART denotes average reaching time.
    }
    \label{tab:nav_results}
    \begin{tabular}{lcc}
        \toprule
        \textbf{Setting} & \textbf{SR (\%)$\uparrow$} & \textbf{ART (s) $\downarrow$} \\
        \midrule
        Full QuadVerse (Sim) & 92 & 13.49 \\
        \textbf{Full QuadVerse (Real)} & \textbf{84} & \textbf{14.24} \\
        QuadVerse w/o Reconstruction & 36 & 19.36 \\
        QuadVerse w/o Residual Comp. & 76 & 18.20 \\
        \bottomrule
    \end{tabular}
    \vspace{-12pt}
\end{wraptable}

Real-world results are averaged over 25 trials with randomized initial robot poses and goal locations, while simulation results are averaged over 200 episodes.
We compare four settings: \textit{Full QuadVerse (Sim)}, the complete pipeline evaluated in simulation; \textit{Full QuadVerse (Real)}, the same trained policy deployed on hardware; \textit{QuadVerse w/o Reconstruction}, a policy trained without scene-specific real-to-sim reconstruction; and \textit{QuadVerse w/o Residual Comp.}, a policy trained in the reconstructed scene but without residual actuator compensation.

As shown in Table~\ref{tab:nav_results}, the full QuadVerse policy achieves an 84\% success rate on the real robot, approaching its 92\% performance in simulation. 
Removing scene-specific reconstruction causes a large performance drop, suggesting that reconstructed visual and geometric scene structure is important for this navigation task. 
Removing residual actuator compensation also reduces the success rate and increases the reaching time. 
This degradation mainly comes from poorer tracking of high-level velocity commands: the robot accumulates trajectory deviations, follows less direct navigation paths, and is less likely to reach the goal within the episode horizon.
Overall, these results show that the aligned simulation built by QuadVerse supports zero-shot visual-navigation policy deployment.


\section{Conclusion}
\label{sec:conclusion}
We introduced \textbf{QuadVerse}, an integrated real-to-sim-to-real framework for quadruped robot learning in reconstructed scenes. 
By combining batched 3DGS rendering, semantic mesh-based contact calibration, and replay-trained residual actuator compensation, QuadVerse reduces discrepancies across visual perception, physical interaction, and actuator dynamics. 
Experiments demonstrate improved reconstruction quality, contact replay fidelity, and locomotion tracking accuracy, enabling robust zero-shot visual-navigation deployment without task-specific real-world rollouts.


\section{Limitations and Future Work}
\label{sec:limitation}
While QuadVerse provides a practical framework for real-to-sim-to-real transfer, several limitations remain. 
First, our scene representation assumes static geometry and rigid collision meshes, which may be insufficient for strongly deformable terrains, such as shrub-covered areas, sandy ground, or soft soil. 
Second, our contact calibration is most suitable for regions with clear semantic boundaries and relatively homogeneous physical properties; it may be less effective on terrains with high-frequency spatial variations in contact properties. 
In addition, posterior contact calibration and dynamics compensation rely on offline trajectory data collected in the target environment, which may limit scalability to rapidly changing scenes or substantially different robot hardware. 
Future work will explore dynamic scene representations, finer-grained contact calibration, deformable terrain models, and online adaptation to reduce environment-specific calibration requirements.

\clearpage


\bibliography{references}  

\clearpage
\input{chapters/8_appendix}

\end{document}